%% file: root.tex
\pdfoutput=1 
\documentclass[letterpaper, 10 pt, conference]{ieeeconf}  
\usepackage{amsmath}
\usepackage{adjustbox}
\usepackage{graphicx,float}
\usepackage{epstopdf}
\usepackage{caption}
\usepackage{pdflscape}
\usepackage{amssymb}
\usepackage{amsfonts}
\usepackage{pgfplots}
\usepackage{bbm}
\usepackage{color}
\usepackage{multirow}
\usepackage{tikz}
\usepackage{dblfloatfix}
\usepackage{float}
\usepackage{hyperref}
\usepackage{adjustbox}
\usepackage{balance}
\usepackage{algorithm}
\usepackage{algpseudocode}
\usepackage{relsize}
\newcommand\norm[1]{\lVert#1\rVert}

\IEEEoverridecommandlockouts                              

\title{\LARGE \bf
 RAVE: A Framework for Radar Ego-Velocity Estimation

}

\author{Vlaho-Josip Štironja$^{1}$, Luka Petrović$^{1}$, Juraj Peršić$^{2}$, Ivan Marković$^{1}$, and Ivan Petrović$^{1}$% <-this % stops a space
\thanks{$^{1}$University of Zagreb Faculty of Electrical Engineering and Computing, Laboratory for Autonomous Systems and Mobile Robotics (LAMOR), Unska 3, HR-10000, Zagreb, Croatia
        {\tt\small \{name.surname\}@fer.hr}}%
\thanks{$^{2}$Calirad d.o.o.,
Augusta Harambašića 4, HR-10000, Zagreb, Croatia
        {\tt\small  juraj.persic@calirad.net}}%
}

\begin{document}

\maketitle
\thispagestyle{empty}
\pagestyle{empty}

\begin{abstract}
\input{text/abstract}
\end{abstract}

\input{text/intro_and_related}

\input{text/methodology}
\input{text/results}

\input{text/conclusion}

\section*{ACKNOWLEDGMENT}
This research has been supported by the European Regional Development Fund under the grant PK.1.1.02.0008 (DATACROSS).

\balance
\bibliography{thebibliography}
\bibliographystyle{IEEEtran}

\end{document}

%% file: text/abstract.tex
State estimation is an essential component of autonomous systems, usually relying on sensor fusion that integrates data from cameras, LiDARs and IMUs.
Recently, radars have shown the potential to improve the accuracy and robustness of state estimation and perception, especially in challenging environmental conditions such as adverse weather and low-light scenarios.
In this paper, we present a framework for ego-velocity estimation, which we call RAVE, that relies on 3D automotive radar data and encompasses zero velocity detection, outlier rejection, and velocity estimation.
In addition, we propose a simple filtering method to discard infeasible ego-velocity estimates.
We also conduct a systematic analysis of how different existing outlier rejection techniques and optimization loss functions impact estimation accuracy.
Our evaluation on three open-source datasets demonstrates the effectiveness of the proposed filter and a significant positive impact of RAVE on the odometry accuracy.
Furthermore, we release an open-source implementation of the proposed framework for radar ego-velocity estimation accompanied with a ROS interface. 

%% file: text/intro_and_related.tex
\section{Introduction}

\begin{figure}[!t]
    \centering
    \includegraphics[width=0.38\textwidth]{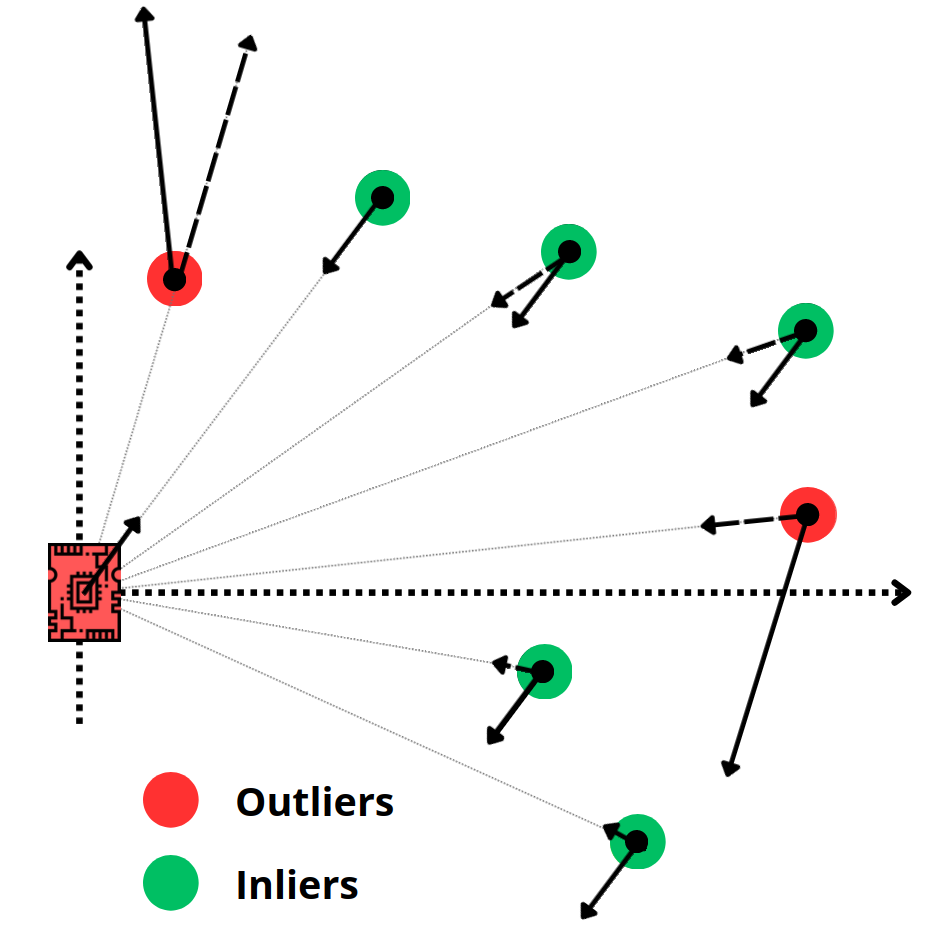}
    \caption{The ego velocity model relies on data from stationary targets (shown in green). Radars measure the relative velocity based on the Doppler effect (indicated by the dashed arrow). Outliers, such as dynamic targets or ghost targets, are indicated by red dots.}
    \label{fig::model}
    \vspace{-5mm}
\end{figure}

Radar sensors are steadily gaining attention in the fields of odometry and simultaneous localization and mapping (SLAM)~\cite{iriom, Michalczyk_closed,kramer_2024, movro} 
because, unlike other commonly used sensors such as cameras and LiDARs, they can operate in adverse environmental conditions such as rain, snow, and direct sunlight
Furthermore, radar sensors are suitable for both indoor and outdoor environments and are less prone to drift than inertial measurement units (IMUs)~\cite{survey_ro_2023}.
Most sensor fusion approaches based on radar data estimate radar ego-velocity separately, and then fuse the velocity within a variant of the extended Kalman filter or a factor graph-based estimator~\cite{ekf_rio_doer,doer_rio_online,4DRadarSLAM,doer_gnss,rvio_doer}.
An alternative approach is to fuse measurements in a tightly-coupled manner within a filter framework \cite{Michalczyk_closed, Michalczyk_multi}.
Both approaches assume that every object in the environment is static, so that it is possible to establish a relationship between the Doppler velocity of the points and the ego velocity of the sensor, as shown in Fig.~\ref{fig::model}.
Both approaches also carry out outlier rejection, which is needed since radars contain dynamic points and ghost targets.

Radar odometry methods leverage radar data in different ways, e.g., by including ego-velocity estimates, using scan registration results, or relying on tracked points which can then constrain the state.
While scan registration and point tracking have reliability issues with 3D low-cost radar systems, ego-velocity estimation can facilitate consistent and accurate odometry estimation~\cite{ekf_rio_doer}.
An instantaneous velocity estimation method based on radar measurements was proposed in \cite{inst_kellner}, using RANSAC for outlier rejection and least squares (LS) for the refinement step.
LS treats the angle of arrival as if it is error-free, while in ~\cite{odr_kellner} orthogonal distance regression (ODR) was used. 
However, it was shown that ODR does not significantly improve the estimation performance compared to LS, while it requires additional computational effort ~\cite{ekf_rio_doer}.
A similar approach was later extended to  3D~\cite{ekf_rio_doer,doer_rio_online,rvio_doer,doer_gnss}, resulting in a method for estimating radar ego-velocity  dubbed~\textit{reve}.

The \textit{reve} method was used in a variety of recent SLAM and radar-based odometry estimation methods \cite{4DRadarSLAM, pgslam_li,DeRo}.
4D iRIOM~\cite{iriom} estimates the ego-velocity using graduated non-convexity (GNC), which is a non-minimal solver typically more robust than RANSAC, albeit being unsuitable for more than $100$ observations.
In \cite{multi_huang}, authors employed Cauchy robust loss kernel, which is deterministic, unlike RANSAC, and more computationally efficient than GNC.
Since the Cauchy robust loss kernel is sensitive to the initial estimate, authors set the initial values using preintegrated body frame velocity from IMU measurements rotated to the radar's frame.
In \cite{kramer_2021} and \cite{stahoviak}, the ego-velocity is estimated using radar and inertial error terms in the cost function, with the novelty being the introduction of maximum likelihood estimation sample consensus (MLESAC) to reject outliers and robust norms to further limit the impact of poor measurements.
Another method presented in \cite{drio}, uses a reliable ground detection algorithm that can jointly estimate radar velocity, performing reliably even in complex dynamic scenarios.

In this paper, we propose a framework for radar ego-velocity estimation, called RAVE, which uses simple radar measurement filtering to discard inaccurate radar ego-velocity estimates without relying on environmental information (e.g., ground plane).
We focus on \textit{automotive radars}, which produces a sparse point cloud of detections, typically represented as a list of range, radial velocity, azimuth, and elevation per detection \cite{survey_ro_2023}.
The proposed filter improves the ego-velocity estimation, leading to better performance in downstream tasks such as odometry and SLAM.
While there are many different methods for estimating radar ego-velocity based on point cloud data, there is a lack of comparison between them.
To address this gap, we compare existing methods and conduct a systematic evaluation of how different outlier rejection techniques and optimization loss functions affect estimation accuracy.
Our comprehensive accuracy comparison includes three open-source datasets, namely Coloradar \cite{coloradar}, IRS \cite{rvio_doer}, and View of Delft (VoD) \cite{vod}.
In addition, we provide an open-source software package Radar Velocity Estimator, named RAVE\footnote{\href{https://bitbucket.org/unizg-fer-lamor/rave}{\tt\small bitbucket.org/unizg-fer-lamor/rave}}, which implements radar ego-velocity estimation framework that features the proposed filter, different outlier rejection techniques and optimization loss functions, and a Robot Operating System (ROS) interface.

%% file: text/methodology.tex
\section{The RAVE framework}

\begin{figure*}[!ht]
    \centering
    \adjustimage{width=0.95\textwidth,center}{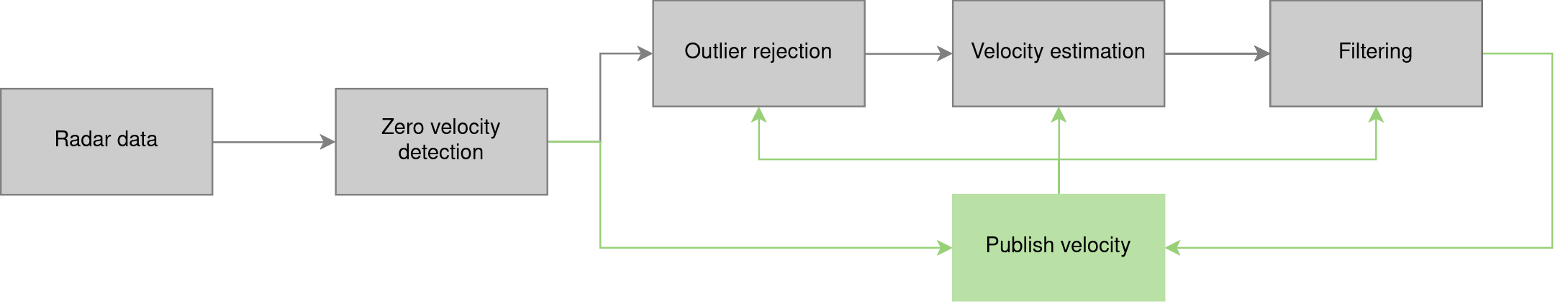}
    \caption{The proposed RAVE framework. Through our filtering method we determine the feasibility of the estimated velocity. The accepted values serve as inputs for the next step, while rejected values are discarded.  }
    \label{fig::rave_pipeline}
\end{figure*}

\subsection{Framework overview}
The proposed RAVE framework is suitable for radar sensors that provide 3D coordinates ($x$,$y$,$z$) and Doppler shift measurements of the target, i.e., relative velocity values.
The ego velocity is estimated using a single scan, a method commonly referred to as an instantaneous method.
Upon arrival of the radar data, we first perform zero-velocity detection similar to \cite{ekf_rio_doer}.
If the median of the Doppler velocity measurements in the point cloud falls below an user-defined minimum threshold and a relatively small percentage of the data (e.g., $25$\%) does not meet this criterion, we conclude that the velocity is zero.

Ego-velocity estimation typically assumes that the majority of radar detections $\mathrm{D}_{i,k}$ ($i \in (1,M)$) within a single scan at a time instant $k$ are static and provide Doppler velocity measurements $\mathrm{v}_{\mathrm{D},k}$ and 3D object positions ${}_{\mathrm{R}} \mathbf{p}_{\mathrm{RD},\mathrm{i},k}$.
Here an expression ${}_{\mathrm{{{{\scalebox{0.56}{A}}}}}}\mathbf{{t}_{{{\scalebox{0.56}{BC}}}}}$ represents a vector from frame $\mathcal{F}_{{\scalebox{0.56}{B}}}$ to frame $\mathcal{F}_{{\scalebox{0.56}{C}}}$ expressed in frame $\mathcal{F}_{{\scalebox{0.56}{A}}}$.
Any valid static radar detection measurement must satisfy the following conditions:
\begin{equation}
    -v_{\mathrm{{{D}}{}}_{\mathrm{i},k}}= \frac{{}_{{{R}}}\mathrm{\mathbf{p}_{{RD}{}}}_{\mathrm{i},k}}{||{}_{{{R}}}\mathrm{\mathbf{p}_{{RD}{}}}_{\mathrm{i},k}||} \cdot  \mathrm{{}_{{R}}}\mathrm{\mathbf{v}}_{{{R},k}},
    \label{vd}
\end{equation}
where $\mathrm{{}_{{R}}}\mathrm{\mathbf{v}}_{{{R},k}}$ is the 3D translational radar velocity at the time instant $k$.
Hence, with at least three non-coplanar raw radar measurements, it is possible to compute the 3D translational radar velocity \cite{barron20053d}.
In practice, radar point clouds often contain outliers in the form of non-stationary objects, ghost targets or alike.
Similarly to the state-of-the-art methods, we distinguish the process of outlier rejection from the velocity estimation (using inliers), which constitutes a separate, modular part of the proposed framework.

Once inliers are determined, we construct a custom loss function, offering the flexibility to use different robust loss functions such as Cauchy, Huber, and General norm robust loss function presented in \cite{barron}, from which all other loss functions (e.g., German-McClure, L1-L2, L2, Welsch/Leclerc) can be derived.
The simplest form of a general loss function is defined as follows:
\begin{equation}
    f(x,\alpha,c)= \frac{| \alpha -2 |}{\alpha}  \biggl( \Bigl(\frac{(x/c)^2}{|\alpha -2 |}+1\Bigl)^{\frac{\alpha}{2}} -1 \biggl),
\end{equation}
where $\alpha \in {\rm I\!R} $ represents shape parameters controlling robustness, and $c>0$ is a scale parameter regulating the size of the quadratic bowl near $x=0$ \cite{barron}.
The introduction of a robust loss function can significantly reduce the effects of outliers that lead to errors in an estimate.
After constructing the loss function, we solve the optimization problem using the Broyden–Fletcher–Goldfarb–Shanno (BFGS) algorithm \cite{bfgs}.
The obtained velocity estimates are then filtered using the proposed filter, which is described in the following section.
The proposed framework is illustrated in Fig.~\ref{fig::rave_pipeline}.

\subsection{The proposed filter}
Since the velocity estimator at times produces unlikely velocity estimates, we implement a simplified filter to check the estimated velocity feasibility. 
The reason for this is that we have encountered situations where outlier rejection methods had difficulties in correctly identifying outliers, especially in scenarios with numerous moving objects, leading to unlikely velocity estimates. 

For each new estimate, we apply the sliding window technique and calculate the average velocity norm based on the last $N$ estimates.
The estimate is not kept if the difference between the mean and the newly estimated velocity norm is larger than a pre-specified threshold $T_1$, while the difference between the last and the current estimate is also too large, implying an acceleration above a threshold $a_{max}$.
It is crucial to choose an appropriate value for $N$ to ensure that it represents relatively recent values, while the thresholds $T_1$ and $a_{max}$ should not be set too low, as this could lead to the rejection of accurately estimated values during fast movements. 
Using a larger window size requires a higher $T_1$ value as the moving average would be ``slower''.
It is also important to adjust the $N$ and $T_1$ parameters according to the sampling frequency of the radar, as a lower sampling frequency for the same $N$ requires a higher $T_1$ value than a higher sampling frequency.
This approach is intended to filter out highly unlikely estimates, which are rare but can significantly degrade the accuracy in downstream tasks, such as odometry and SLAM.

When filling up an empty queue, we first check only the difference between the last and the current estimate. 
In our implementation, we do not consider the possibility of incorrect estimates at the first measurement, since most sequences start in a steady state where zero velocity detection provides reliable estimates. 
Finally, the correctly estimated velocity values that have successfully passed our filter can serve as initial values for the subsequent steps.
This is especially important when using the robust Cauchy kernel loss, which is sensitive to initial values. The proposed filtering algorithm is summarized in Algorithm \ref{algo}.

\begin{algorithm}
\caption{The RAVE filter}
\begin{algorithmic}[1]
\Statex \textbf{Inputs:} Sliding windows size $N$, velocity estimates $\mathbf{v}_{k-(N+1):k}$, thresholds $T_1$,  $a_{max}$
\Statex \textbf{Output:} Updated window of last $N$ valid velocity estimates

\State Calculate the average norm of velocity estimates in the window $\mathrm{n}_{\text{avg}} = \frac{1}{N} \sum_{i=1}^N \norm{\mathbf{v}_{k-i}}$

%\State 
\If{  $\lvert \mathrm{n}_{\text{avg}} - \norm{\mathbf{v}_{k}}\rvert < T_1$ \, and \, $\norm{\mathbf{v}_k -\mathbf{v}_{k-1}} / (t_k - t_{k-1}) < a_{max}$}
    \State Estimated velocity is valid: drop $\mathbf{v}_{k-(N+1)}$ and add $\mathbf{v}_{k}$ to the window
\Else 
    \State Estimated velocity is invalid: Reject $\mathbf{v}_{k}$
\EndIf
\end{algorithmic}
\label{algo}
\end{algorithm}

\subsection{Implementation details}

Due to the modular nature of the proposed framework, there are several parameters that should be chosen depending on the context of the problem at hand. 
In our implementation, we have explicitly specified some of these  parameters for which we have empirically found that they lead to accurate velocity estimation for commonly used radars.
We set the threshold for zero velocity detection to 0.05\,m/s.
For the proposed sliding-window filter, we chose the size of the filter queue $N=5$, the norm difference threshold $T_1=7.5$\,m/s, and the maximum translational acceleration $a_{max}=10$\,m/s$^2$.

For outlier rejection, most existing radar ego-velocity estimation methods use a 3-point RANSAC approach \cite{ekf_rio_doer,doer_rio_online}.
RAVE also implements RANSAC, but it additionally implements GNC and MLESAC outlier rejection methods.
Furthermore, our implementation supports multiple loss functions.
RAVE allows outlier rejection and velocity estimation to be combined into a unified step by using the robust Cauchy loss function for the velocity estimation loss, similar to \cite{multi_huang}.
If the inliers are identified using RANSAC or MLESAC, the velocity estimation loss function can be generated using least squares (LS), truncated LS (TLS), weighted LS (WLS) with the signal-to-noise ration used for weighting, weighted truncated LS (WTLS), or any other custom optimization loss functions tailored to a particular problem.
The choice of outlier rejection method and optimization loss function and their advantages and disadvantages are analyzed and discussed in the following section. 

%% file: text/results.tex
\section{EXPERIMENTAL RESULTS}
\label{experimental_results}

In this section, we present results of the experimental validation of different ego-velocity estimators based on radar measurements. 
First, we investigate the influence of the proposed filter on the results using sequences from the IRS dataset.
Second, we compare different outlier rejection methods (RANSAC, MLESAC, GNC) and estimation algorithms (LS, TLS, Cauchy robust loss) to investigate the advantages and disadvantages of each method.
Finally, we demonstrate the importance of ego-velocity estimation for accurate odometry using MOVRO \cite{movro} as the odometry algorithm.

\subsection{Datasets}
We evaluate the performance of RAVE on three open-source datasets: the IRS dataset, the ColoRadar dataset, and the VoD dataset.

The IRS dataset is the Radar Thermal Visual Inertial Dataset, which includes a wide range of indoor and outdoor sequences. 
It consists of a powerful IMU (Analog Devices ADIS16448), an FCMW radar (Texas Instruments IWR6843AOP) and a monocular grayscale camera (IDS UI-3241). 
The sensors are synchronized using hardware triggers and synchronization signals via a microcontroller board. 
The ground truth odometry data for some sequences are provided by the Vicon motion capture system (MoCap), while the ground truth data for the remaining sequences are analyzed using VINS \cite{vins} with loop closures.

ColoRadar, the 3D millimeter wave radar dataset, contains more than two hours of data collected in a large indoor and outdoor environment using a handheld sensor rig for robotic mapping and state estimation. 
This rig is equipped with two FMCW radar sensors (TI MMWACSRF-EVM, TI AWR1843BOOST-EVM), a 3D lidar (Ouster OS1) and an IMU (Lord Microstrain 3DM-GX5-25). 
The high-precision ground truth is achieved either with Vicon poses or with a globally optimized pose graph with IMU, lidar and loop closure constraints using \cite{coloradar_slam}.

The View-of-Delft (VoD) dataset \cite{vod} is a vehicle dataset containing 8693 frames of synchronized and calibrated 64-layer LiDAR (Velodyne HDL-64 S3), a (stereo) camera pair (1936×1216 px) and 3+1D radar data (ZF FR-Gen21 3+1D) acquired in complex urban traffic, as well as odometry data estimated using RTK-GPS, IMU and wheel odometry. 
The dataset is synchronized so that the LiDAR serves as the primary sensor, with the timestamps of the nearest camera and radar data rewritten to match the LiDAR timestamps. 
This synchronization method poses a challenge in distinguishing poses for velocity estimation. 
Therefore, we decided to perform a comparison with the MOVRO \cite{movro} odometry framework using different radar ego-velocity estimation methods.

For the IRS dataset, our evaluation is specifically focused on  Mocap sequences, as they contain the ground truth velocity parameters derived from the Vicon motion capture system, from which we obtain the ground truth translation velocity parameters in the radar's frame using the Euler's equation. 
Euler's equation for rigid bodies calculates velocity $\mathbf{v}_{\mathrm{a}}$ from $\mathbf{v}_{\mathrm{b}}$ using the following equation:
\begin{equation}
\mathbf{v}_{\mathrm{a}}=\mathbf{v}_{\mathrm{b}}+ \boldsymbol{\omega}_{\mathrm{b}} \times \mathbf{p}_{\mathrm{ba}},
\end{equation}
where $\boldsymbol{\omega}_{\mathrm{b}}$ and $\mathbf{p}_{\mathrm{ba}}$ denote the angular velocity of a rigid body around point b and the position vector from point b to point a, respectively. 
Due to negligible translational distance between the ground truth origin and the radar frame within the IRS dataset, we opted to neglect $\boldsymbol{\omega} \times \mathbf{p}_{\mathrm{ab}}$ part which simplifies ground truth velocity calculation. 

For ColoRadar sequences, ground truth calculation we followed the same procedure as in \cite{4DEgo}. 
To obtain ground truth ego-velocity data in time of estimates, we use linear interpolation.

\subsection{Effectivness of the proposed filter}
\begin{figure}[t]
    \centering
    \includegraphics[width=0.5\textwidth]{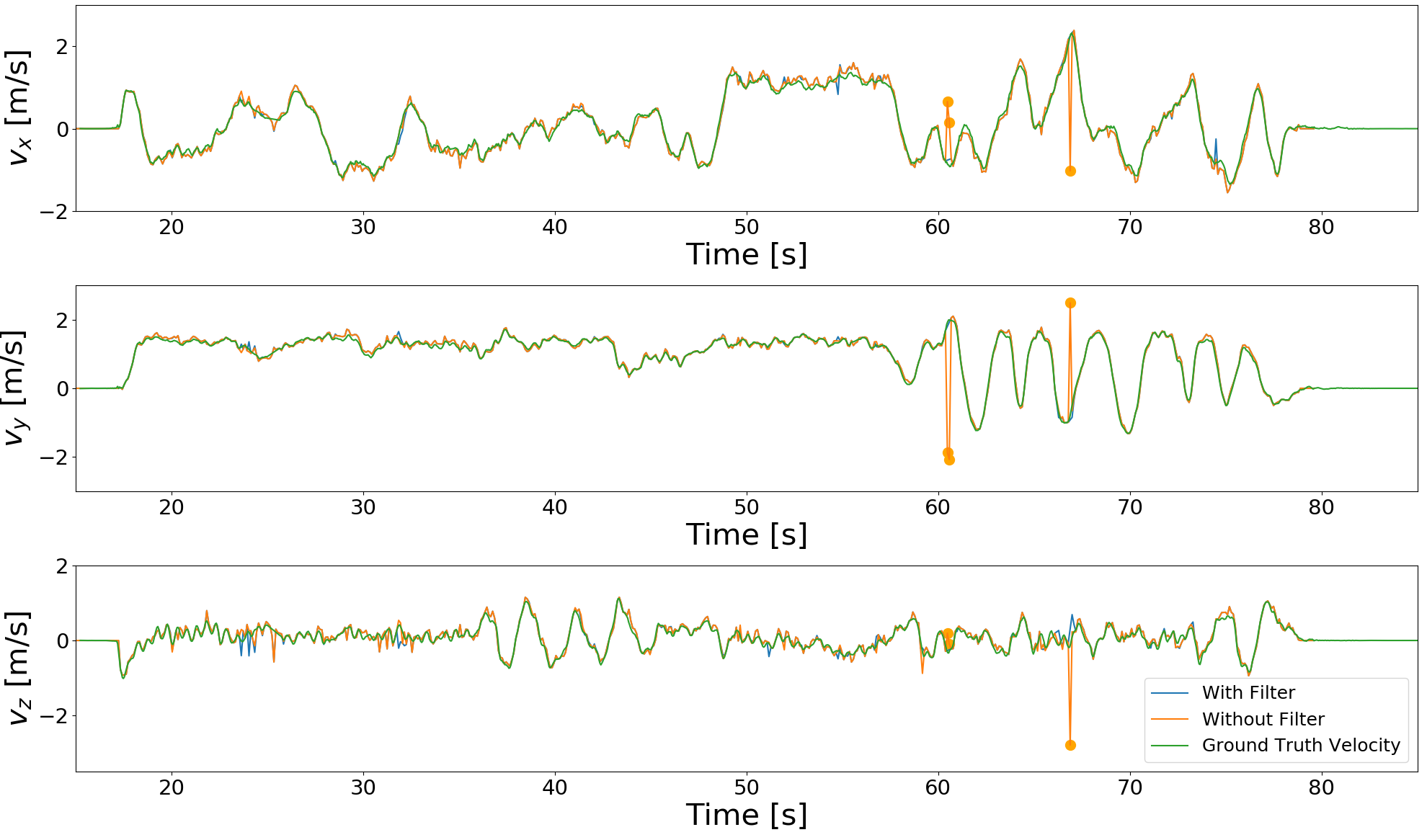}
    \caption{ Visualization of velocity estimation with and without the proposed filter on the Mocap dark fast sequence. }
    \label{fig::v}
\end{figure}

To evaluate the effect of the filter described in Algorithm~\ref{algo}, we estimate the ego-velocity of the radar on three IRS sequences with Mocap ground truth. 
We use a common approach with RANSAC for outlier rejections and LS as loss function, both with and without the proposed filter. 
The obtained results are shown in Table~\ref{irs_f_1}. Outliers have a larger impact on the root mean square error (RMSE) than on the average velocity error (AVE). 
Overall, our filter reduced both errors, especially in difficult scenarios (e.g. Mocap difficult), while similar results were obtained in other cases. 
The Mocap easy sequence does not contain any sudden motion changes, which explains the same error value with and without the filter. 
A visualization of the velocity estimation for the Mocap dark fast sequence, shown in Fig.~\eqref{fig::v}, illustrates the positive effect of the filter. 
In this particular case, our filter identified and discarded 3 outliers in the ego-velocity estimates out of a total of 789 ego-velocity estimates. 
Due to the performance improvement of the proposed filter, it will be used for the remaining experiments in RAVE to mitigate the impact of outliers.

\renewcommand{\tabcolsep}{2.8pt}
\begin{table}[t]
\centering
\renewcommand{\arraystretch}{1.5}
\begin{adjustbox}{width=\columnwidth,center}
\begin{tabular}{c  c  c  c  c  c  c  c  c  c  c }
\hline
& \multirow{2}{*}{Filter}  & \multicolumn{3}{c}{Mocap easy}  & \multicolumn{3}{c}{Mocap dark fast} & \multicolumn{3}{c}{Mocap difficult}\\
\cline{3-11}
&  & $v_x$ & $v_y$ & $v_z$ & $v_x$ & $v_y$ & $v_z$ & $v_x$ & $v_y$ & $v_z$ \\
\hline 
\multirow{2}{*}{AVE} & No & \textbf{0.044} & \textbf{0.027} & 0.047 & 0.087 & 0.069 & 0.091 & 0.118 & 0.083 & 0.104  \\
& Yes & \textbf{0.044} & \textbf{0.027} & \textbf{0.046} & \textbf{0.074} & \textbf{0.050} & \textbf{0.077} & \textbf{0.104} & \textbf{0.060} & \textbf{0.082}  \\
\hline
\multirow{2}{*}{RMSE} & No & \textbf{0.060} & \textbf{0.035} & 0.071 & 0.224 & 0.271 & 0.281 & 0.255 & 0.265 & 0.242  \\
& Yes  & \textbf{0.060}  & \textbf{0.035} & \textbf{0.069} & \textbf{0.098} & \textbf{0.065} & \textbf{0.115} & \textbf{0.151} & \textbf{0.082} & \textbf{0.121} \\
\hline
\end{tabular}
\end{adjustbox}
\caption{AVE [m/s] and RMSE [m/s] for three Mocap IRS dataset sequences with and without the proposed filter.}
\label{irs_f_1}
\end{table}

\subsection{Analysis of outlier rejection techniques and loss functions}

\renewcommand{\tabcolsep}{5pt}
\begin{table*}[t]
\renewcommand{\arraystretch}{1.3}
\noindent\makebox[\textwidth]{
\begin{tabular}{  c  c c  c  c  c c  c c  c c  c c  c  c c  c c  }
 \hline
 \multirow{2}{*}{Method}   & \multicolumn{3}{c}{Mocap easy}  & \multicolumn{3}{c}{Mocap medium}  &  \multicolumn{3}{c}{Mocap dark}  & \multicolumn{3}{c}{Mocap dark fast}  & \multicolumn{3}{c}{Mocap difficult}   \\  \cline{2-16}
   &   $v_x$ & $v_y$  & $v_z$ & $v_x$    & $v_y$ &$v_z$  & $v_x$ & $v_y$    & $v_z$ & $v_x$    & $v_y$ & $v_z$   & $v_x$ & $v_y$   & $v_z$      \\
 \hline
  RANSAC + LS &  0.060 &  \textbf{0.035}& 0.069& 0.133 & 0.069 & 0.096 &  0.137 & 0.079 & 0.114 & \textbf{0.096}  & \textbf{0.066} &  0.116 & 0.136 &  0.081 &  \textbf{0.119} \\
  MLESAC + LS & 0.067  & 0.040 & 0.080 &  0.141 & 0.069 & 0.134 & 0.138 &  0.083 & 0.135 & 0.217 & 0.126 & 0.147 & 0.162  &  0.090&  0.157 \\
  RANSAC + TLS & 0.059 &  0.036 & 0.072 & 0.135 &  0.070 & 0.093 & 0.144 &  0.087 & 0.118 & 0.101 & 0.082 &  0.119 &  0.174 &  0.090&  0.125  \\
  MLESAC + TLS & 0.069   & 0.043 & 0.083 &  0.142 & 0.071 & 0.117 & 0.148 &  0.097 & 0.132 & 0.181 & 0.121 & 0.151 &0.193 & 0.140 & 0.172  \\
  RANSAC + Cauchy & 0.060  & \textbf{0.035}  & 0.069 & \textbf{0.130} & 0.069 & \textbf{0.092} &  0.142 &  0.085 & \textbf{0.111} & 0.098 & 0.080 &  \textbf{0.112} & 0.166  & 0.086 & 0.121  \\
 MLESAC + Cauchy & 0.067 & 0.042 & 0.080 & 0.137 & 0.079 & 0.117 &  0.146  & 0.092 &  0.135 & 0.209 & 0.135 & 0.151 & 0.237 & 0.150 & 0.164 \\
        GNC & 0.109 &  0.070 & 0.135 & 0.150 & 0.104  & 0.164 &  0.166 &  0.124& 0.171 & 0.277 & 0.178 & 0.217 & 0.258 & 0.245 &  0.242 \\
    Cauchy & \textbf{0.058} & \textbf{0.035}  & \textbf{0.068} &  \textbf{0.130} & \textbf{0.068} &  \textbf{0.092} & \textbf{0.134} & \textbf{0.078} &  0.112 & 0.097 & \textbf{0.066} &  0.113 & \textbf{0.133} & \textbf{0.081} &  0.122 \\
     Huber & 0.062 & 0.036 & 0.080 & 0.134 & 0.070 & 0.116 &0.135  & 0.082  & 0.127 & 0.104 &  0.068  & 0.127 & 0.144 & 0.084 &  0.147 \\
     \hline
\end{tabular}}
\caption{RMSE [m/s] for Mocap IRS dataset sequences. The best results for each dataset sequence are highlighted in bold, and all values are rounded to three decimal places.}
\label{irs}
\end{table*}

Next, we performed a comprehensive analysis of outlier rejection and loss function methods on IRS dataset sequences with Mocap ground truth (Table~\eqref{irs}) and on three ColoRadar sequences using a low-cost single-chip radar (Table~\eqref{coloradar_}).
We compared RANSAC and MLESAC as outlier rejection techniques, using LS, TLS, and robust Cauchy norm to construct the optimization loss function.
We also evaluated the performance of GNC, Cauchy, and Huber robust norms without outlier rejection.

\begin{figure}[t]
    \centering
    \includegraphics[width=0.5\textwidth]{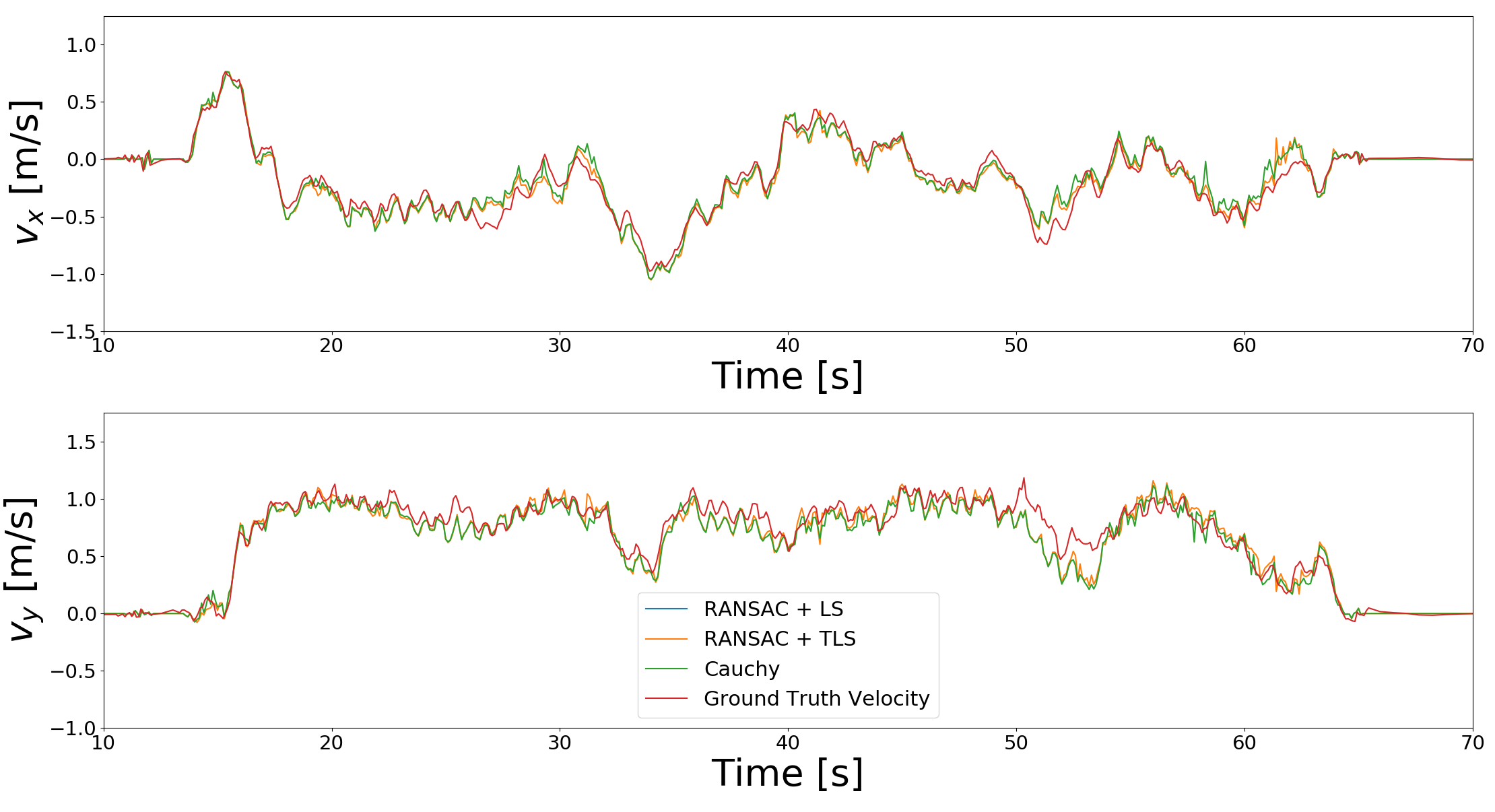}
    \caption{Visualization of velocity estimation with three different methods on the aspen\_run0 sequence.}
    \label{fig::coloradar_v}
\end{figure}

As expected, the accuracy of ego-velocity estimation depends on the difficulty of the sequence, i.e., the Mocap difficult and Mocap dark fast sequences have larger errors than the Mocap easy sequence of the IRS dataset. 
From the results obtained, it can be concluded that the robust Cauchy kernel loss provides the best performance in velocity estimation. 
For the sequences of the IRS dataset, it is more accurate without RANSAC or MLESAC, while for the sequences of the Coloradar dataset, the combination with RANSAC provides the most accurate results, albeit with a longer runtime of the algorithm compared to the other methods. 
It is important to note that these differences can be attributed to the larger point cloud in the Coloradar dataset, which contains  higher number of outliers. 
Furthermore, the TLS loss does not show a significant advantage over the simple LS. 
A similar conclusion was drawn in \cite{ekf_rio_doer}, where the authors found that orthogonal distance refinement does not provide significant improvements over LS. 
The GNC method does not provide competitive results compared to RANSAC in combination with LS, possibly due to high noise in the low-cost radar data. 
Fig.~\eqref{fig::coloradar_v} shows comparison of the velocity estimation for three different methods (RANSAC+LS, RANSAC+TLS, Cauchy) with the smallest RMSE on the aspen\_run0 sequence. 
The lowest ego-velocity accuracy is observed in the z-axis, which is due to the lower resolution of the radar at elevation angles compared to azimuth angles.

\renewcommand{\tabcolsep}{2.7pt}
\begin{table}[t]
\centering
\renewcommand{\arraystretch}{1.5}
\begin{tabular}{ c c  c c  c  c  c  c  c  c }
\hline
\multirow{2}{*}{Method} & \multicolumn{3}{c}{aspen\_run0} & \multicolumn{3}{c}{arpg\_lab\_run0} & \multicolumn{3}{c}{ec\_hallways\_run0} \\
\cline{2-10}
& $v_x$ & $v_y$ & $v_z$ & $v_x$ & $v_y$ & $v_z$ & $v_x$ & $v_y$ & $v_z$ \\
\hline
R + LS & 0.061 &  \textbf{0.085} & 0.174 &  0.090 & 0.117 & 0.225 & 0.135 & 0.159 & 0.249 \\
M + LS  & 0.079  & 0.087 & 0.219 & 0.151 & 0.143 & 0.336 & 0.171 & 0.176 & 0.310 \\
R + TLS & 0.061 &  \textbf{0.085} & 0.175 &  0.091 & \textbf{0.116} & 0.225 & 0.134 & 0.167 & \textbf{0.244} \\
M + TLS  & 0.085 & 0.088  & 0.225 & 0.147 & 0.148 & 0.336 & 0.182 & 0.184 & 0.320 \\
R + C  & \textbf{0.060} & \textbf{0.085}  & \textbf{0.173} & \textbf{0.089} & \textbf{0.116} & \textbf{0.221} & \textbf{0.130} & \textbf{0.156} & 0.249 \\
M + C  & 0.084 & 0.088  & 0.219 & 0.131 & 0.141 & 0.313 & 0.169 & 0.178 & 0.280 \\
GNC  & 0.114 & 0.139 & 0.228 & 0.148 & 0.164 & 0.385 & 0.193 & 0.204 & 0.362 \\
C  & 0.064 & \textbf{0.085} & 0.196 & 0.093 & 0.116 & 0.247 & 0.154 & 0.182 & 0.268 \\
H  & 0.078 & 0.092 & 0.226 & 0.118 & 0.136 & 0.336 & 0.163 & 0.181 & 0.317 \\
\hline
\end{tabular}
\caption{RMSE [m/s] for three ColoRadar dataset sequences.  The best results of each dataset sequence are highlighted in bold and all values are rounded to three decimal places. The symbols R, M, C, and H stand for the RANSAC, MLESAC, Cauchy, and Huber robust norms, respectively.}
\label{coloradar_}
\end{table}

\subsection{Impact on odometry accuracy}
\begin{figure}[t]
    \centering
    \includegraphics[width=0.5\textwidth]{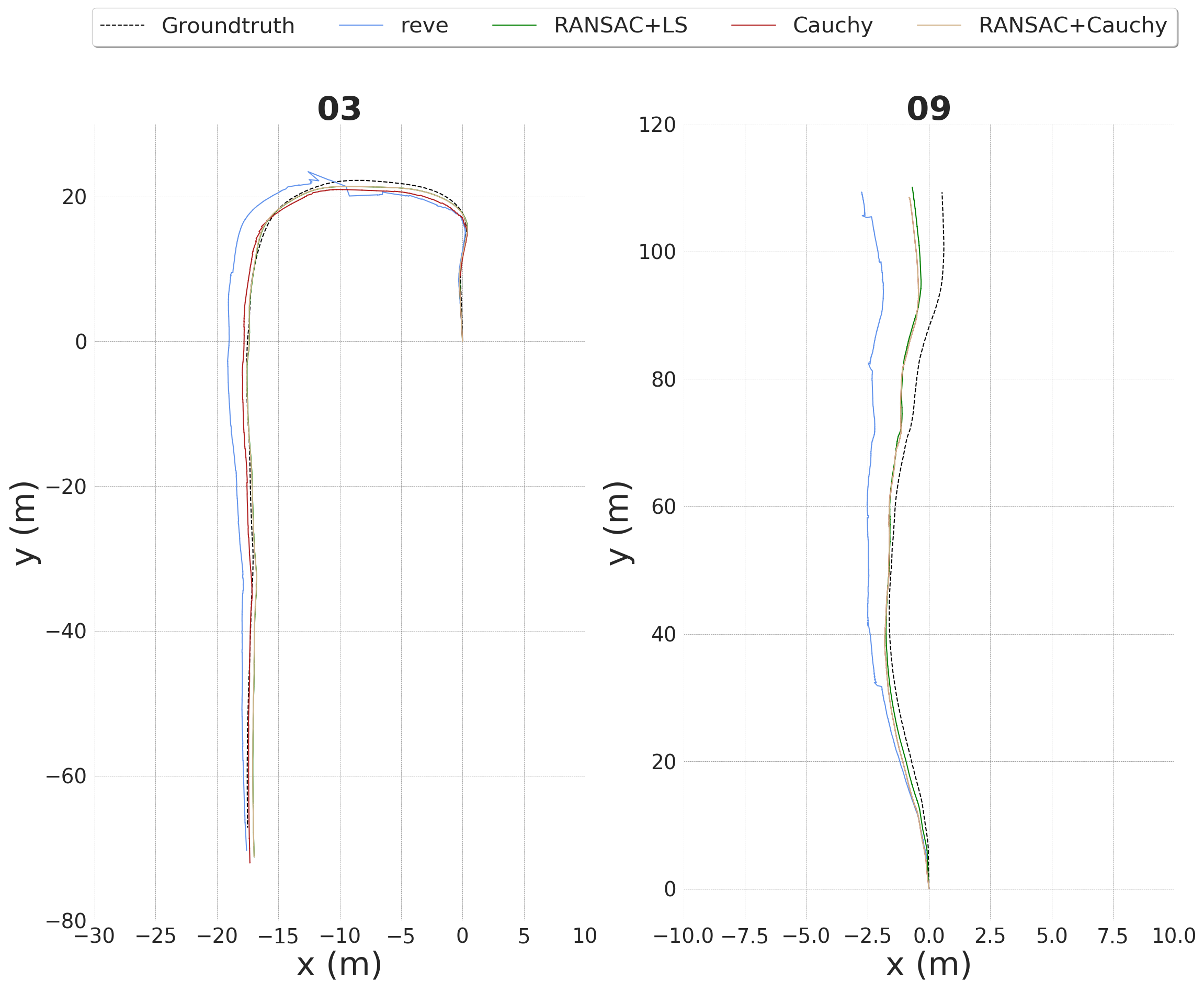}
    \caption{MOVRO \cite{movro} trajectories on the VoD dataset for four different approaches to radar ego velocity estimation.}
    \label{fig::vod}
\end{figure}

We evaluate the RAVE velocity estimation within the MOVRO \cite{movro} odometry framework on the VoD dataset. 
The relative pose error (RPE), i.e., the relative translation error $t_{rel}$ and rotation error $r_{rel}$, is used to measure the performance. 
MOVRO integrates radar ego velocity measurements with monocular odometry pose data. 
Monocular odometry poses are analyzed using ov2slam ~\cite{ov2slam}, while four different methods were used for the radar velocity measurements: the \textit{reve} package ~\cite{ekf_rio_doer} and our implementation of RANSAC+LS, RANSAC+Cauchy, and the Cauchy method. 
The results obtained for three VoD sequences are shown in Table.~\ref{table_vod}, while the visualization of the trajectories can be seen in Fig.~\ref{fig::vod}. 
We can see the benefit of improved ego-velocity accuracy and discarded estimation outliers on odometry accuracy, especially on  $t_{rel}$, where the proposed ego-velocity estimation method outperforms a state-of-the-art ego-velocity estimation method with each tested combination of outlier rejection and optimization loss functions~\cite{ekf_rio_doer}.

\begin{table}[t]
\centering
\renewcommand{\arraystretch}{1.5}
\begin{tabular}{ c  c  c  c  c  c  c  }
\hline
\multirow{2}{*}{Method} & \multicolumn{2}{c}{03} & \multicolumn{2}{c}{09} & \multicolumn{2}{c}{22} \\
\cline{2-7}
& $t_{rel}$ & $r_{rel}$ & $t_{rel}$ & $r_{rel}$ & $t_{rel}$ & $r_{rel}$ \\
\hline
\textit{reve}~\cite{ekf_rio_doer} & 0.1190 & 0.2696  & 0.1100 & 0.1755 & 0.0883 & 0.2543  \\
RANSAC + LS & 0.0844 & 0.2455  & 0.0845 & \textbf{0.1575} & \textbf{0.0588} & 0.2885  \\
RANSAC + Cauchy & \textbf{0.0841} & \textbf{0.2429} & \textbf{0.0844} & \textbf{0.1575} & 0.0579 & 0.2435\\
Cauchy & 0.0859 & 0.2596  &0.0993 & \textbf{0.1575}& 0.0658 & \textbf{0.2382}  \\
\hline
\end{tabular}
\caption{Relative translation and relative rotation error ( $t_{rel}$ [m],  $r_{rel}$ [deg]) for MOVRO \cite{movro} with four different approaches to estimate ego-velocity using VoD dataset sequences, rounded to three decimal places. }
\label{table_vod}
\end{table}

%% file: text/conclusion.tex
\section{Conclusion}

In this paper we have presented the RAVE framework for ego-velocity estimation that relies on 3D automotive radar data. 
We propose to use a simple filtering method to discard infeasible ego-velocity estimates. 
In addition, we provide a comprehensive comparative analysis of how different outlier rejection methods and optimization loss functions affect the accuracy of radar ego-velocity estimation. %
We also provide an open-source implementation of the proposed framework with various outlier rejection and optimization loss methods, as well as a ROS interface. 
We evaluated the performance of the RAVE framework on three different datasets. 
Our experiments demonstrated the effectiveness of the proposed  filter for ego-velocity estimation and a positive effect of the proposed framework on odometry accuracy, with RAVE improving odometry performance in comparison to a state-of-the-art method~\cite{ekf_rio_doer}.